\documentclass{article}

\PassOptionsToPackage{numbers, compress}{natbib}

\usepackage[final]{neurips_2024}

\usepackage[backref=page,breaklinks,colorlinks]{hyperref}
\hypersetup{
     citecolor    = magenta,
     urlcolor = blue,
     linkcolor = blue,
}
\usepackage[utf8]{inputenc} 
\usepackage[T1]{fontenc}    
\usepackage{url}            
\usepackage{booktabs}       
\usepackage{tabularx}
\usepackage{amsfonts}       
\usepackage{amsmath}
\usepackage{mathtools}
\usepackage{nicefrac}       
\usepackage{microtype}      
\usepackage{xcolor}         
\usepackage{graphicx}
\usepackage{subfigure}
\usepackage{comment}

\title{Which bits went where? Past and future transfer entropy decomposition with the information bottleneck}

\author{Kieran A. Murphy\\
Dept. of Bioengineering, University of Pennsylvania\\
\texttt{kieranm@seas.upenn.edu}\\
\And 
Zhuowen Yin\\
Dept. of Bioengineering, University of Pennsylvania\\
\texttt{zhuowen@seas.upenn.edu}\\
\And 
Dani S. Bassett \\
Dept. of Bioengineering, University of Pennsylvania \\
Dept. of Electrical \& Systems Engineering, University of Pennsylvania \\
Dept. of Neurology, Perelman School of Medicine, University of Pennsylvania \\
Dept. of Psychiatry, Perelman School of Medicine, University of Pennsylvania\\
Dept. of Physics \& Astronomy, University of Pennsylvania\\
The Santa Fe Institute\\
Montreal Neurological Institute, McGill University\\
\texttt{dsb@seas.upenn.edu}
}

\begin{document}

\maketitle

\begin{abstract}
Whether the system under study is a shoal of fish, a collection of neurons, or a set of interacting atmospheric and oceanic processes, transfer entropy measures the flow of information between time series and can detect possible causal relationships.
Much like mutual information, transfer entropy is generally reported as a single value summarizing an amount of shared variation, yet a more fine-grained accounting might illuminate much about the processes under study. 
Here we propose to decompose transfer entropy and localize the bits of variation on both sides of information flow: that of the originating process's past and that of the receiving process's future.
We employ the information bottleneck (IB) to compress the time series and identify the transferred entropy.
We apply our method to decompose the transfer entropy in several synthetic recurrent processes and an experimental mouse dataset of concurrent behavioral and neural activity. 
Our approach highlights the nuanced dynamics within information flow, laying a foundation for future explorations into the intricate interplay of temporal processes in complex systems.

\end{abstract}

\vspace{-4.5mm}
\section{Introduction}
\vspace{-2mm}
Causality forms the backbone of explanations in science.
Detecting cause and effect from observational data alone is generally impossible, as there can always be unobserved shared causes.
However, certain signatures have been identified that can indicate \textit{possible} causality, such as Granger causality~\cite{granger1969investigating}, transfer entropy~\cite{schreiber2000transfer}, and directed information~\cite{massey1990causality}.
Measuring such signatures can be a powerful step toward understanding a complex system, with applications as broad as groups of fish~\cite{butail2016model}, neural signals~\cite{novelli2021inferring, staniek2008symbolic}, and Earth-scale climate processes~\cite{runge2019inferring}.

Transfer entropy quantifies the additional information a source process shares with a target process after accounting for the information contained in the target process's history~\cite{schreiber2000transfer,james2016critique}.
As a conditional mutual information, transfer entropy is nonparametric and generalizes to arbitrary probability distributions and relationships between variables, in contrast to Granger causality; transfer entropy and Granger causality are equivalent for Gaussian variables~\cite{barnett2009granger}.
On the other hand, conditional mutual information can be challenging to estimate from limited data samples~\cite{mukherjee2020cmi,staniek2008symbolic}.

Consider modeling the movement of the blue fish in Fig.~\ref{fig:schematic}a so as to forecast its future given its past.
If the red fish causes the blue fish to alter its behavior, then the accuracy of the forecast will improve when information about the past of the red fish is incorporated into the model.
The converse is not necessarily true---that an improvement in accuracy implies a causal relationship---but quantifying information flow in this way is often the closest to causality one can get with observational data alone (i.e., without having the ability to intervene in the data collection process)~\cite{pearl2009causality,barnett2009granger}.

While the above framing in terms of forecasting with and without the source's past is the common way to present transfer entropy, there is an equivalent alternative that we will leverage in this work.
Again employing the fish of Fig.~\ref{fig:schematic}a, consider the task of inferring the red fish's movement given the blue fish's movement over the same time frame.
If the accuracy of the model improves when incorporating the future of the blue fish, there is information flow from the past of the red fish to the future of the blue fish.
With these two equivalent expressions for transfer entropy, we can focus on the effect of incorporating either the past of the red fish or the future of the blue fish into a predictive model.
The net information change in either case must be equal to the transfer entropy, but the originating and terminating variation can look different on either side of the information flow.
Our goal in this work is to identify the variation on either side by using the information bottleneck.

The information bottleneck (IB) is a method to isolate targeted variation in a learned compression scheme~\cite{tishbyIB2000,alemiVIB2016}, allowing for control over learned representations~\cite{goldfeld2020IBReview} and fine-grained inspection of \textit{what} the information is~\cite{dib_iclr}.
Here we propose an IB scheme to encapsulate the transfer entropy from either side of an information flow into a learned compression space.
The encapsulated information can then be decomposed in terms of multiple sources via the distributed IB~\cite{dib_iclr,dib_pnas}, inspected as a local quantity varying in time~\cite{lizier2008local}, and used as a targeted representation for downstream tasks~\cite{kalajdzievski2023teb}.

\vspace{-1mm}
\section{Approach}
\begin{figure*}
    \centering
    \includegraphics[width=\linewidth]{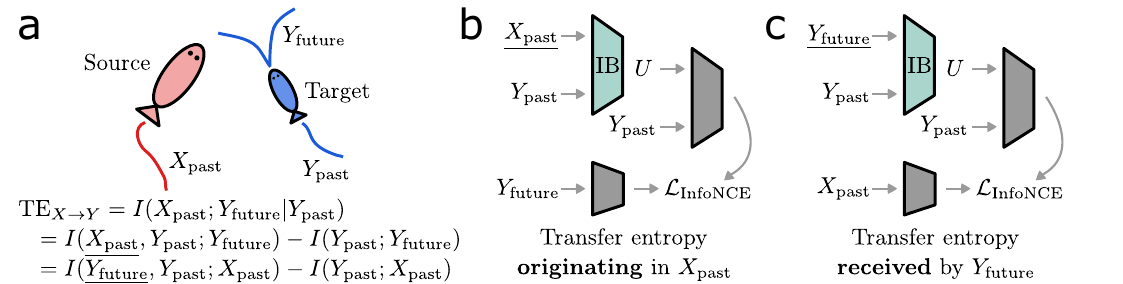}
    \caption{\textbf{Localizing transfer entropy in the past and future.}
    \textbf{(a)} If the past of a large red fish (the source) helps predict the future of a small blue fish (the target) after accounting for its past, then there is positive transfer entropy.
    Equivalently, if the future of the blue fish helps infer the past of the red fish after accounting for the blue fish's past, then there is positive transfer entropy.
    \textbf{(b)} A machine learning scheme to extract the transfer entropy from the source's past, using the information bottleneck (IB).
    \textbf{(c)} An analogous scheme to extract the transfer entropy from the target's future.
    }
    \label{fig:schematic}
\end{figure*}

\vspace{-2mm}
Transfer entropy measures information flow from a \textit{source} time series to a \textit{target} time series and is the information gained in either of two equivalent scenarios (Fig.~\ref{fig:schematic}a): \textbf{(i)} the information that the source's past adds to the target's past about the target's future, and \textbf{(ii)} the information that the target's future adds to the target's past about the source's past.
We will set up both scenarios as a constrained communication problem with the information bottleneck~\cite{tishbyIB2000} in order to identify and then decompose the bits of transferred entropy in the source's past and in the target's future.

Let $X_t$ and $Y_t$ represent two random processes, the source and target, and let $t$ be a discrete index for time.
We assume the processes are stationary and will use the same time horizon, $\tau$, into the past and future.
For shorthand, $Y_{-\tau:0}$ and $Y_{0:\tau}$ (using \texttt{python} indexing for $\tau$ steps into the past and future, respectively~\cite{james2016critique}) will be written as $Y_\text{past}$ and $Y_\text{future}$, respectively.

The forecasting information inherent to the target process is the mutual information~\cite{cover1999elements} between the recent past and the near future, $I(Y_\text{past};Y_\text{future})$.
The transfer entropy from source to target is the increase in forecasting information when incorporating the recent history of the source,

\begin{equation}\label{eqn:te_past}
    \text{TE}_{X\rightarrow Y}=I(X_\text{past},Y_\text{past};Y_\text{future}) - I(Y_\text{past};Y_\text{future}).
\end{equation}

Equivalently, the transfer entropy can be found in the target's future as the amount of information added about the source's past,

\begin{equation}\label{eqn:te_future}
    \text{TE}_{X\rightarrow Y}=I(Y_\text{future},Y_\text{past};X_\text{past}) - I(Y_\text{past};X_\text{past}).
\end{equation}

Eqns.~\ref{eqn:te_past} and \ref{eqn:te_future} can be seen as the difference in predictive power between two extremes of a continuous spectrum parameterized by the amount of information incorporated about $X_\text{past}$ and $Y_\text{future}$, respectively.
We will traverse the spectrum by compressing the added piece of information with the information bottleneck (Fig.~\ref{fig:schematic}b).
Importantly, to compress the added information requires $Y_\text{past}$ as context in both cases, which can simply be passed along as an additional input to the encoder.
Letting $U=f(X_\text{past},Y_\text{past})$ be the compression of the additional information, and $\beta$ be an information cost, minimizing the following Lagrangian with respect to $U$ traverses the spectrum relevant to Eqn.~\ref{eqn:te_past},
\begin{equation}\label{eqn:te_ib}
    \mathcal{L}=\beta I(X_\text{past},Y_\text{past};U) - I(U,Y_\text{past}; Y_\text{future}).
\end{equation}
In the limit $\beta \rightarrow \infty$, no information is used about the source and we recover the self-forecasting information $I(Y_\text{past};Y_\text{future})$.
At the other extreme, $\beta \rightarrow 0$, all information from the source is used.  
The difference in the loss between these extremes is the transfer entropy.
In practice, we decrease $\beta$ logarithmically over the course of a single training run to obtain the full spectrum of transferred information; the transfer entropy is obtained by the difference between the endpoints of the trajectory. 
To optimize with machine learning, we employ the same information cost as a $\beta$-VAE~\cite{betavae,alemiVIB2016}, and a lower bound on the forecasting information via InfoNCE~\cite{oord2018InfoNCE},
\begin{equation}\label{eqn:mlteb}
    \mathcal{L} = \beta \ \mathbb{E}[D_\text{KL}(p(u|x_\text{past},y_\text{past}) || r(u))] - \mathbb{E} [\mathcal{L}_\text{NCE}(f(u,y_\text{past}), g(y_\text{future}))].
\end{equation}
The first expectation is over $p(x_\text{past},y_\text{past})$ and the second is over $p(u|x_\text{past},y_\text{past})p(x_\text{past},y_\text{past},y_\text{future})$.
$r(u)$ is a prior distribution, taken to be the standard normal $\mathcal{N}(\boldsymbol{0}, \boldsymbol{1})$, and $f(\cdot)$ and $g(\cdot)$ are functions parameterized by neural networks.
While Eqn.~\ref{eqn:mlteb} targets the transfer entropy in $X_\text{past}$, the transfer entropy in $Y_\text{future}$ is found analogously with the compression variable $U=f(Y_\text{future},Y_\text{past})$ (Fig.~\ref{fig:schematic}c).

Without loss of generality, assume the source and target are compositions of processes---e.g., multiple physiological signals from each fish in Fig.~\ref{fig:schematic}a---so that $X_t=(X_t^1,...X_t^N)$.
For added interpretability around the extracted information, we compress each component process and/or each point in time separately by distributing an information bottleneck to each as a separate random variable~\cite{dib_iclr,dib_pnas}. 
The Lagrangian corresponding to Eqn.~\ref{eqn:te_ib} is
\begin{equation}
    \mathcal{L}=\beta\sum_i^N \left[I(X^i_\text{past},Y_\text{past};U^i)\right] - I(\boldsymbol{U},Y_\text{past}; Y_\text{future}).
\end{equation}\label{eqn:tedib}

\begin{figure*}
    \centering
    \includegraphics[width=\linewidth]{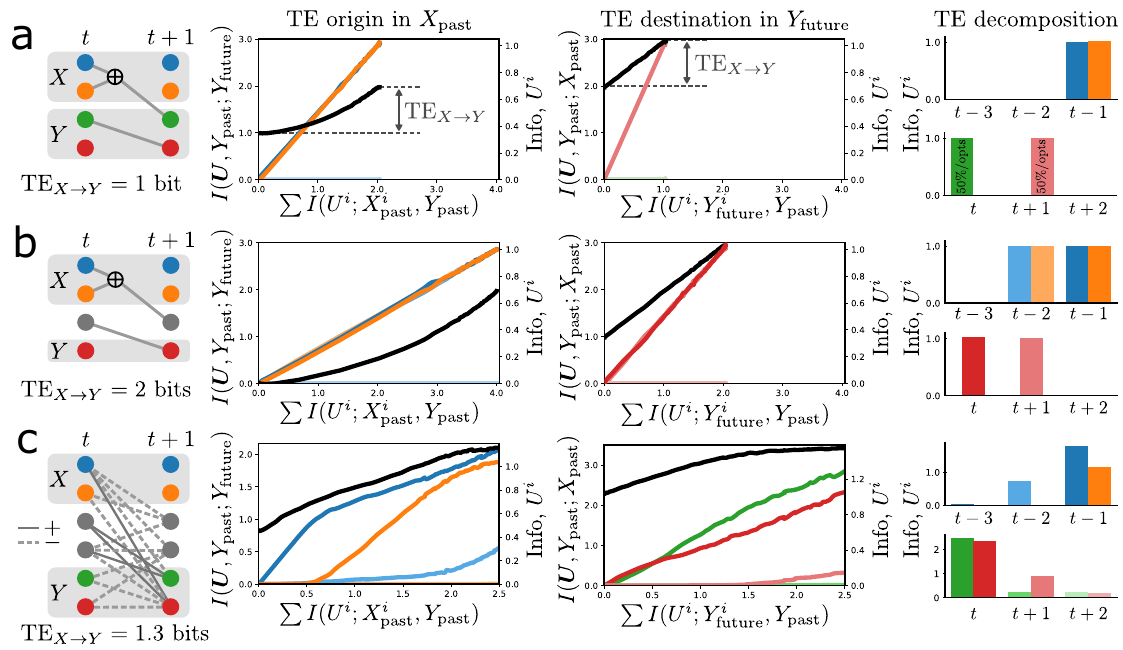}
    \caption{\textbf{Transferred entropy in binary-valued recurrent networks.}
    \textbf{(a)} \textit{Left:} The update rule for four processes, where nodes without inputs (blue, orange) are randomly sampled at each timestep.
    The source and target processes are indicated by the shaded boxes and marked $X$ and $Y$, respectively.
    \textit{Middle:} Distributed information planes that visualize the decomposition of transfer entropy in the source's past and the target's future.
    \textit{Right:} The share of transfer entropy residing in different timesteps of the source's past (top) and target's future (bottom) (taken from the rightmost point of the trajectories in the middle).
    \textbf{(b)} Same as panel \textbf{a}, but with different target processes.
    \textbf{(c)} Same as panel \textbf{a}, but with randomly generated connection weights and an integrate-and-fire scheme.
    }
    \label{fig:synthetic}
\end{figure*}

We note important distinctions from the recently proposed Transfer Entropy Bottleneck~\cite{kalajdzievski2023teb}: we propose a simpler joint encoding strategy, a decomposition of the future transfer entropy, and a distributed IB scheme for more granularity when analyzing information flow between processes.

\vspace{-1mm}
\section{Results and Discussion}
\vspace{-2mm}
We will now work through synthetic examples to build intuition for the transfer entropy decomposition, and then analyze an experimental dataset from a group of mice where neural activity and behavioral data were collected simultaneously.
All quantities are in bits unless specified otherwise.
Implementation specifics can be found in the Appendix.

\textbf{Synthetic Boolean networks.} 
For the systems of binary-valued processes in Fig.~\ref{fig:synthetic}, we generated trajectories of length $10^4$ and trained the end-to-end setups shown in Fig.~\ref{fig:schematic}b\&c with a time horizon $\tau=3$.
A classic example regarding transfer entropy~\cite{james2016critique} (slightly modified) is shown in Fig.~\ref{fig:synthetic}a\&b.
Blue and orange are random processes that determine the next timestep of the green process through an \texttt{XOR} interaction, and the red process stores the most recent state of green.
There is positive transfer entropy when considering the blue and orange processes together as a single composite source, and either or both of the green and red processes as targets.
With the distributed IB, we bottlenecked each process and each timestep to identify the origin and the terminus of the transferred entropy.

The difference in total information $I(\boldsymbol{U},Y_\text{past};Y_\text{future})$ between the start and end of optimization is the transfer entropy, and therefore the same when focusing on the transfer entropy's origin in $X_\text{past}$ and its destination in $Y_\text{future}$.
However, the trajectories are qualitatively different for the past and future decompositions due to the nature of interactions between the constituent processes~\cite{dib_pnas}.
In Fig.~\ref{fig:synthetic}a, the single bit of transfer entropy starts as two bits in the blue and orange sources' past at timestep $t-1$, and then ends as one bit in the future of the red process at timestep $t+1$.
Note that the green and red process contain the same information (with a delay of one), meaning the transferred entropy equivalently resides in the green process at timestep $t$.
When repeating the distributed IB optimization for Fig.~\ref{fig:synthetic}a, the transferred entropy was localized in the green process at timestep $t$ in half of the optimizations and in the red process at timestep $t+1$ in the other half.
The degeneracy resolves if we ignore the green process (Fig.~\ref{fig:synthetic}b), and the transfer entropy increases to two bits.
Four bits originate in the blue and orange processes and then terminate in the red process over the future two timesteps.

The benefit of machine learning becomes more apparent when decomposing information flow in more complicated processes.
In Fig.~\ref{fig:synthetic}c, a set of connected processes includes two sources (blue and orange, random), two hidden states (grey), and two targets (green and red).
Nodes with connections from the previous timestep follow an integrate-and-fire update rule, meaning they will fire if the sum of their inputs, weighted by +1 or -1 as indicated by the solid and dashed lines, is greater than zero.
The share of information from different timesteps showcases the nontrivial recurrence established by the randomly selected connection scheme and provides clues about the 
underlying process interactions.

\begin{figure*}
    \centering
    \includegraphics[width=\linewidth]{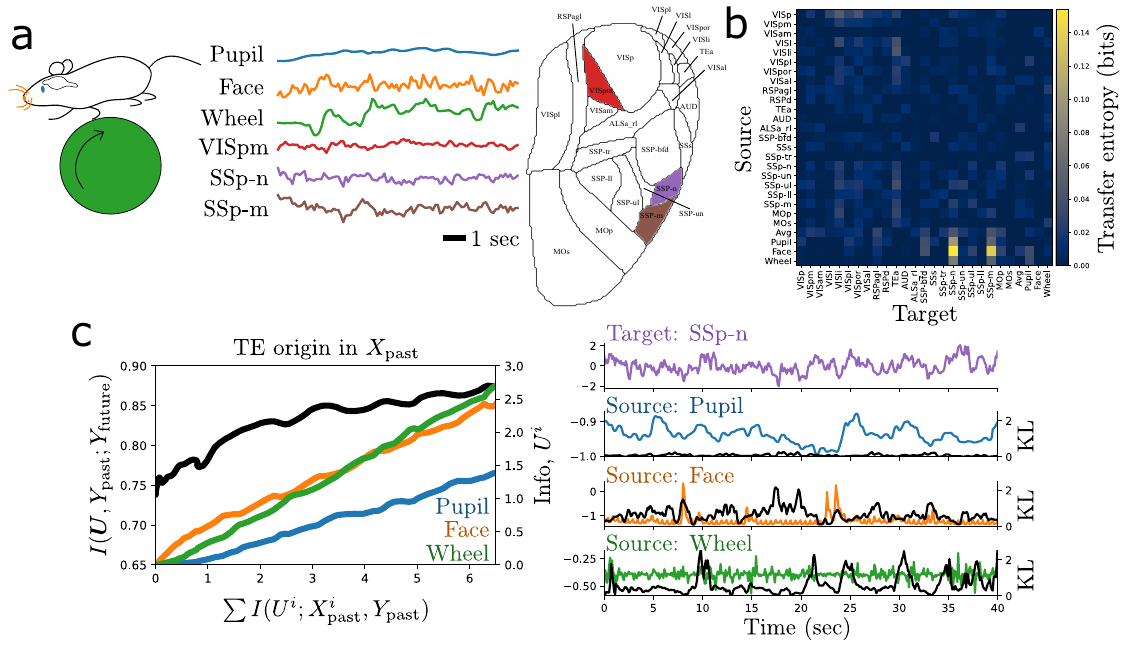}
    \caption{\textbf{Transfer entropy between brain and behavior.}
    \textbf{(a)} Concurrent neural and behavioral recordings were taken of six mice; example time series shown on the right with the brain regions shown with matching colors in the atlas.
    \textbf{(b)} Pairwise transfer entropy between the 23 brain regions, the reference average signal (Avg), and three behavioral streams.
    \textbf{(c)} Transfer entropy decomposition from behaviors to the purple region from \textbf{a}, the primary somatosensory area for the nose (SSp-n).
    The instantaneous Kullback-Leibler (KL) cost in natural units (nats) per channel (black) is shown concurrently with the raw time series (colored).
    }
    \label{fig:mouse}
\end{figure*}

\textbf{Experimental mouse data.}
We next analyzed calcium imaging and spontaneous behavioral data from six mice publicly uploaded with \citet{benisty2024rapid}, with a time horizon of 1 second (10 samples).
A rough schematic of the experimental setup is shown in Fig.~\ref{fig:mouse}a.

We estimated the pairwise transfer entropy between all 23 regions, their common average signal (`Avg'), and the three behavioral time series, one pair at a time using the difference in forecasting InfoNCE values with and without the source (Fig.~\ref{fig:mouse}b).
The largest pairwise values link behavior to brain region SSp-n, and we decompose the multivariate transfer entropy in Fig.~\ref{fig:mouse}c.
The pupil time series contained the second largest transfer entropy with SSp-n (Fig.~\ref{fig:mouse}b), though its contribution was the smallest when combined with the face and wheel data, suggesting redundant 
information.

At any point along the information bottleneck spectrum, we can inspect the learned compression scheme.
We encoded a stream of the validation data with the three learned encoders at an intermediate value of $I(\boldsymbol{U},Y_\text{past}; Y_\text{future})$ and display the Kullback-Leibler (KL) cost---whose expectation serves as a penalty on the transmitted information in Eqn.~\ref{eqn:mlteb}---at each point in time (Fig.~\ref{fig:mouse}c, right).  
Note each encoding scheme embeds the source past along with the target past as context, meaning the spikes in KL cost could arise from an interaction between the two signals.
In this way we obtain a fine-grained picture suggesting where in the source time series the transfer entropy spawns, which could be paired with a local estimate of the transfer entropy~\cite{martinez2020can} to take a microscope to information transfer between processes.

\textbf{Discussion.} The quantification of information flows between parts of a system via transfer entropy is an important step toward a deeper understanding and serves as a screening for possible causation.
In this work, we localized transfer entropy on both sides of an information flow: from its origin in the source's past to its terminus in the target's future.
We note that although both routes to transfer entropy outlined are equivalent, the prediction task involved is qualitatively different.
When compressing the source's past, the prediction task is to forecast the target's future from its past.
By contrast, when compressing the target's future, the prediction task is to link the target's past with the source's past, which will generally share a lower amount of information than the target's past with its future.
The difficulty of optimization for the two formulations may thus be different in practice.

Finally, we note that conditioning on additional processes when computing transfer entropy allows one to exclude the influence of the processes~\cite{schreiber2000transfer,lizier2008local}.
The proposed framework readily handles such additional processes by appending them to both instances of $Y_\text{past}$ in the schematics of Fig.~\ref{fig:schematic}b, c.


\newpage
\appendix
\section{Implementation specifics}

All experiments were implemented in Tensorflow and run on a single computer with a 12 GB GeForce RTX 3060 GPU.
An iPython notebook to run the synthetic examples (Fig. 2) is included with the submission.
Training hyperparameters and architecture details are shown in Tables 1 and 2.

\begin{table}
\centering
\begin{tabularx}{\linewidth}{@{}  *2{>{\raggedright\arraybackslash}X} @{}} 
 \hline
 Parameter & Value \\ 
 \hline\hline
 Bottleneck MLP architecture & [64 dense \texttt{leaky\_ReLU}]\\
 Bottleneck embedding space dimension & 8 \\
 $Y_\text{past}$ encoder architecture & [128 LSTM \texttt{tanh}, 64 dense \texttt{leaky\_ReLU}]\\
 $Y_\text{past}$ embedding space dimension & 32 \\
 Encoder MLP architecture (to shared embedding space) & [256 dense \texttt{leaky\_ReLU}, 256 dense \texttt{leaky\_ReLU}] \\
 Predicted quantity encoder architecture & [128 LSTM \texttt{tanh}, 64 dense \texttt{leaky\_ReLU}]\\
 InfoNCE similarity metric $s(u,v)$ & Euclidean squared\\
 \hline 
 Batch size & 128 \\
 Optimizer & Adam \\
 Learning rate & $3 \times 10^{-4}$ \\
 $\beta_\textnormal{initial}$ & $5 \times 10^{-5}$ \\
 $\beta_\textnormal{final}$ & $3$ \\ 
 Annealing steps & $2 \times 10^4$\\
 Horizon length & 3\\
 \hline
\end{tabularx}
\caption{Training parameters for synthetic experiments.}
\label{tab:hparams1}
\end{table}

\begin{table}
\centering
\begin{tabularx}{\linewidth}{@{}  *2{>{\raggedright\arraybackslash}X} @{}} 
 \hline
 Parameter & Value \\ 
 \hline\hline
 Bottleneck MLP architecture & [256 dense \texttt{leaky\_ReLU}]\\
 Bottleneck embedding space dimension & 8 \\
 $Y_\text{past}$ encoder architecture & [128 LSTM \texttt{tanh}, 128 LSTM \texttt{tanh}, 256 dense \texttt{leaky\_ReLU}]\\
 $Y_\text{past}$ embedding space dimension & 32 \\
 Encoder MLP architecture (to shared embedding space) & [256 dense \texttt{leaky\_ReLU}] \\
 Predicted quantity encoder architecture &[128 LSTM \texttt{tanh}, 128 LSTM \texttt{tanh}, 256 dense \texttt{leaky\_ReLU}]\\
 InfoNCE similarity metric $s(u,v)$ & Euclidean squared\\
 \hline 
 Batch size & 128 \\
 Optimizer & Adam \\
 Learning rate & $3 \times 10^{-4}$ \\
 $\beta_\textnormal{initial}$ & $10^{-3}$ \\
 $\beta_\textnormal{final}$ & $1$ \\ 
 Annealing steps & $2 \times 10^4$\\
 Horizon length & 10\\
 \hline
\end{tabularx}
\caption{Training parameters for mouse data analysis.}
\label{tab:hparams2}
\end{table}

\end{document}